\pdfoutput=1

\documentclass[11pt]{article}

\usepackage[final]{acl}
\usepackage{hhline}
\usepackage[linewidth=1pt]{mdframed}
\usepackage{tabularx}
\usepackage{array}
\usepackage{times}
\usepackage{latexsym}
\usepackage{booktabs}
\usepackage[T1]{fontenc}

\usepackage[utf8]{inputenc}
\definecolor{darkgreen}{RGB}{0, 150, 0} 

\usepackage{microtype}
\usepackage[utf8]{inputenc}
\usepackage{array}
\usepackage{longtable}
\usepackage{inconsolata}
\usepackage{arydshln}

\usepackage{amssymb}
\usepackage{amsmath}
\usepackage{graphicx}
\usepackage{comment}
\usepackage{booktabs}
\usepackage{adjustbox}
\usepackage{multirow}
\usepackage{makecell}
\usepackage{algorithm}
\usepackage{algpseudocode}

\title{\textit{Align-SLM}: Textless Spoken Language Models with Reinforcement Learning from AI Feedback}

\author{Guan-Ting Lin$^{1,2}$\thanks{Work done during internship with Amazon AGI.}\quad Prashanth Gurunath Shivakumar$^{1}$\quad Aditya Gourav$^{1}$ \\ \textbf{Yile Gu$^{1}$ \quad Ankur Gandhe$^{1}$\quad Hung-yi Lee$^{2}$\quad Ivan Bulyko$^{1}$}\\ \\
  $^{1}$Amazon AGI, USA \\
  $^{2}$Graduate Institute of Communication Engineering, National Taiwan University, Taiwan\\
   \\
}
\begin{document}
%
\maketitle
\begin{abstract}
While textless Spoken Language Models (SLMs) have shown potential in end-to-end speech-to-speech modeling, they still lag behind text-based Large Language Models (LLMs) in terms of semantic coherence and relevance. This work introduces the \textbf{Align-SLM} framework, which leverages preference optimization inspired by Reinforcement Learning with AI Feedback (RLAIF) to enhance the semantic understanding of SLMs. Our approach generates multiple speech continuations from a given prompt and uses semantic metrics to create preference data for Direct Preference Optimization (DPO). We evaluate the framework using ZeroSpeech 2021 benchmarks for lexical and syntactic modeling, the spoken version of the StoryCloze dataset for semantic coherence, and other speech generation metrics, including the GPT4-o score and human evaluation. Experimental results show that our method achieves state-of-the-art performance for SLMs on most benchmarks, highlighting the importance of preference optimization to improve the semantics of SLMs.
\end{abstract}
%
%

\section{Introduction}
\label{sec:intro}
Significant strides have been made in Large Language Models (LLMs) by training decoder-only transformer models on vast amounts of text data. In speech processing, Textless NLP~\citep{gslm, pgslm, dglsm, dual} employs discrete speech units to train Spoken Language Models (SLMs) through next speech unit prediction. This approach is particularly promising, as SLMs are end-to-end speech-to-speech models that bypass the traditional cascaded pipeline of Automatic Speech Recognition (ASR) and Text-to-Speech (TTS) systems, enabling joint optimization and real-time human-computer interaction. Furthermore, SLMs are applicable to all spoken languages, including those without written scripts, as they only require unlabeled speech data, thus promoting inclusivity in speech technology.

Despite increasing efforts to develop and improve SLMs—through text model initialization~\citep{twist, gsqa}, speech tokenizer design~\citep{gslm, twist, syllableLM}, text \& speech token interleaving~\citep{sutlm, spiritlm}, scaling data and model~\citep{twist, scaling}— a substantial gap remains between the understanding capabilities of text-based LLMs and SLMs.
Current SLMs, when prompted, often produce speech continuations characterized by repetitive phrases, grammatical inaccuracies, and low relevance. \citet{speechgpt, spectron} propose predicting text during intermediate decoding steps in a chain that mimics the ASR, LM, and TTS tasks within a single model. While these intermediate text steps improve the semantics of the generated speech, they still rely on text tokens as conditions to guide speech generation, and the additional decoding steps introduce latency, preventing real-time interactive SLMs. The question of \textit{\textbf{whether textless SLMs can generate semantically relevant speech}} remains under-explored.

Most research on SLMs has relied \textit{exclusively on next-speech-token prediction}. Few studies have explored alternative optimization objectives. Compared to text subwords, which on average carry more information, speech tokens are finer-grained and less compact. We argue that the next-speech-token prediction task may overlook long-term semantics, as loosely compressed speech units exhibit significant variability along spectral and temporal dimensions. Consequently, SLMs require a better training objective to effectively capture long-range semantics.

Our motivation stems from the observation that SLMs produce inconsistent results, sometimes generating high-quality speech continuations, while at other times producing suboptimal ones. \textbf{\textit{Can we train SLMs to consistently generate better speech continuations while avoiding failures?}}
Drawing inspiration from Reinforcement Learning with Human Feedback (RLHF) for text LLM alignment~\citep{rlhf, dpo}, we propose \textbf{Align-SLM}, the first framework that enhances the semantics of SLMs through RL. Starting with a pre-trained SLM (the open-sourced TWIST~\citep{twist} model), we generate multiple speech continuations from a given speech prompt. The next step is to create preference data (\texttt{prompt, chosen, rejected}) for preference optimization. Since collecting human preferences by listening is costly and time-consuming, following the concept of Reinforcement Learning from AI Feedback (RLAIF), 
we propose an automatic preference data selection strategy with LLM-guided semantic feedback. After preparing the preference data, Direct Preference Optimization (DPO)~\citep{dpo} is applied to learn from the feedback. Additionally, we couple the proposed technique with curriculum learning and demonstrate further improvements. The proposed framework is \textbf{pure speech-to-speech, data efficient}, and does \textbf{not require text injection}~\citep{spiritlm, sutlm} or \textbf{text-to-speech synthesized speech}~\cite{speechgpt}.

We evaluate the SLM's performance using the sWUGGY and sBLIMP from ZeroSpeech 2021~\citep{zerospeech} for lexical and syntactic modeling, and Spoken-StoryCloze and Topic-StoryCloze~\citep{twist} for textual nuances and continuation coherence. Additionally, we perform generative evaluations for speech continuation using (i) human listening tests and (ii) GPT-4 as a proxy for assessing semantic coherence and relevance. The results show that the proposed method achieves superior performance in semantic understanding and speech generation. The contributions can be summarized as follows:
\begin{itemize}
    \item We propose the first preference optimization framework for textless SLMs, demonstrating that preference optimization is crucial for improving the semantics of SLMs.
    \item We develop an automated preference data selection strategy by designing effective semantic metrics to score preference data pairs.
    \item We couple DPO with curriculum learning by iteratively opting for higher criterion of preference data to further enhance performance.
    \item Align-SLM achieves the \textit{\textbf{state-of-the-art}} performance for end-to-end spoken language models on Zerospeech and StoryCloze benchmark (\textbf{77.9\%} on sWUGGY, \textbf{61.1\%} on S-StoryCloze, and \textbf{86.8\%} on T-StoryCloze) and achieves superior Meaningfulness Mean opinion scores with human evaluations.

\end{itemize}

\begin{figure*}[t]
  \centering
  \includegraphics[width=0.93\linewidth]{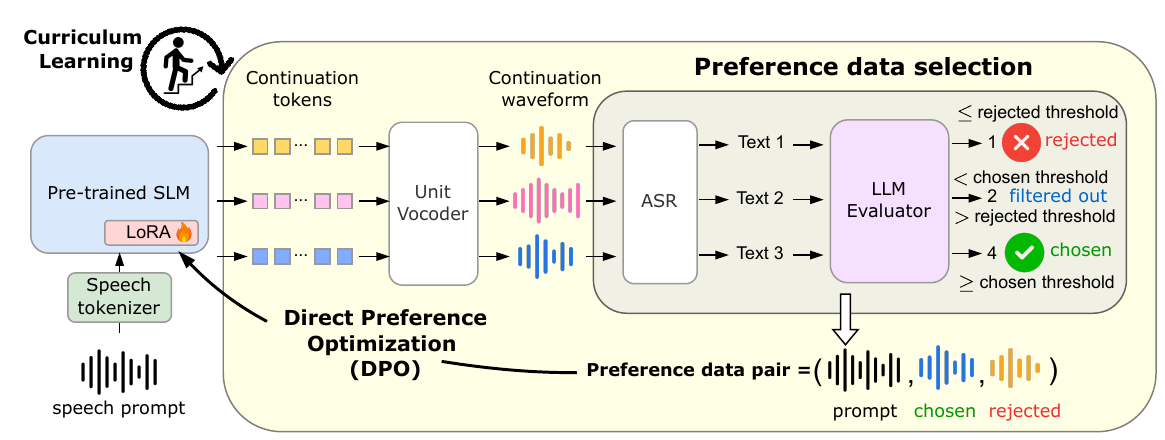}
  \caption{The illustration of the \textbf{Align-SLM} framework. Firstly, tokenize the speech prompt into speech tokens using a speech tokenizer. Then, generate and sample multiple speech continuations from the pre-trained SLM. To create the preference data pairs, the framework uses a unit-based vocoder to synthesize the speech continuations back to waveforms, an ASR model to transcribe the waveforms to text, and an LLM evaluator to rate the quality of the semantics. These preference data pairs are then used for direct preference optimization to train the LoRA adapter in the SLM. This alignment process can be coupled with curriculum learning to further improve performance. }
  \label{fig:framework}
\end{figure*}

\section{Related Work}
\subsection{Spoken Language Models (SLMs)}
Recent advancements in self-supervised representation learning and acoustic unit discovery convert continuous speech signals into discrete speech tokens~\citep{resynthesis}.
SLMs are end-to-end language models with discrete speech tokens, enabling speech continuation given a speech prompt. 
GSLM~\citep{gslm} utilizes speech tokens to train a decoder-only language model and synthesize speech waveforms using a unit-based vocoder. pGSLM~\citep{pgslm} injects prosodic tokens to enhance expressiveness. dGSLM~\citep{dglsm} adopts a dual-tower model for two-channel spoken dialogue modeling.

Although SLMs can generate words and short-term phrases, the long-term semantics of the generated speech are often poor. TWIST~\citep{twist} proposes using a text-based LLM as initialization with large-scale training data to improve semantics. 
VoxtLM~\citep{voxtlm} leverages both paired and unpaired speech in addition to text data for joint training of ASR, TTS, SLM, and Text LM. 
Another research direction is text token prediction as an intermediate step like
SpeechGPT~\citep{speechgpt} and SPECTRON~\citep{spectron}, which perform a chain of tasks (ASR $\rightarrow$ LM $\rightarrow$ TTS) in a single model. 
SUTLM~\citep{sutlm} and SPIRIT-LM~\citep{spiritlm} utilize phrase-level interleaving for speech and text tokens.
Concurrent works like LLaMA-Omni~\citep{llama-omni}, Mini-Omni~\citep{mini-omni}, and Moshi~\citep{moshi} leverage simultaneous text token prediction to guide the speech generation. SyllableLM~\citep{syllableLM} recently propose to use syllable-level coarse speech tokens to improve SLM semantics. 
Compared to prior works focusing on multi-tasking in a single model or using text tokens to guide speech generation, this is the first work to \textit{improve the long-term semantics of speech-only SLMs through preference optimization}.

\subsection{Preference Optimization}
Training the language model with \textit{next-token prediction} is effective for learning human knowledge, but this objective might be different from the human preference. 
RLHF~\citep{rlhf-first, rlhf} leverages an external reward model, combined with proximal policy optimization~\citep{ppo}, to align LLMs based on human feedback. DPO~\citep{dpo} proposes that LLMs can learn implicit rewards, allowing them to perform preference optimization independently, without the need for an external model.
RLAIF~\citep{constitutional, rlaif} demonstrates that using AI as an alternative to human feedback can reduce labor costs and is more scalable, offering performance comparable to RLHF.

These methods show effectiveness in aligning an LLM with human preference. 
However, \textit{there are limited studies on leveraging preference optimization in the speech and audio processing field}. Recently, \citet{speechalign, tts-rlhf} adopted preference optimization for the Text-to-Speech (TTS) model to align the quality of speech synthesis with human preference, \textit{but not for enhancing SLMs' semantics}. \citet{baton, tango2} leverage preference optimization for text-to-audio generation with diffusion model, but the text-to-audio task is very different compared to SLM. 

\section{Align-SLM Framework}
The illustration of the proposed framework is shown in Figure \ref{fig:framework}. 

\subsection{Spoken Language Models}
\label{sec:slm}
The pre-trained SLM used in this work is TWIST~\citep{twist}, a decoder-only transformer model that is trained on the next speech token prediction task with text model initialization. We utilize the TWIST model in two sizes (1.3B and 7B parameters) from the official release\footnote{https://github.com/facebookresearch/textlesslib/tree/main/\\examples/twist (MIT license)}. 
Specifically, the \textbf{speech tokenizer} consists of a self-supervised speech model~\citep{superb, superb-prosody} and K-means clustering. 
In this work, HuBERT~\citep{hubert} is used and the cluster number K is set to 500. 
Notably, when continuous representations are clustered into discrete units, they primarily capture content information, which can be leveraged for modeling and understanding~\citep{disentangle, dual, unit_slu}.
This process first extracts 25Hz frame-level continuous representations from the 11-$th$ layer of the HuBERT model, assigns each frame to its closest cluster index, and then de-duplicates consecutive identical indices to shorten the sequence.
The \textbf{unit-based vocoder} is a HifiGAN-based~\citep{hifigan} model that can convert the discrete units back into a continuous waveform. We use the model checkpoint from the textlesslib~\citep{textlesslib} library.

\subsection{Automatic Preference Data Selection}
\label{sec:preference}
To prepare the preference data pair (prompt, chosen, rejected), given the speech prompt $x$, the nucleus sampling~\citep{nucleus} is used to generate $N$ different continuations $y_1, y_2, ..., y_N$. Ideally, humans can listen to the samples and select the desirable and semantically correct one as the chosen continuation $y_c$, and the semantically incorrect one as the rejected continuation $y_r$. However, it is costly and time-consuming for human annotators to listen to the samples. Following the idea of RLAIF~\cite{constitutional, rlaif} to simulate human feedback, we propose an automatic preference data selection strategy to create preference data pairs.
Since the focus is the semantics of SLMs, which is the content information in the speech, we first use \texttt{Whisper-large-v2}~\citep{whisper} to transcribe the speech into text, then measure the semantics of the transcribed text. In this work, we explore the two types of AI feedback from a text LLM. The text LLM is the open-sourced Mistral 7B (\texttt{instruct-v02})\footnote{https://huggingface.co/mistralai/Mistral-7B-Instruct-v0.2}.

\subsubsection{Continuation Likelihood: Perplexity}
Perplexity (PPL) is a common metric to measure the likelihood of a sentence given a pre-trained language model. In this work, PPL is calculated on the generated transcribed text conditioned on the ground truth text prompt. PPL is used in previous SLM works to evaluate the generation~\citep{gslm, twist}. However, \citet{gslm} found out that SLMs sometimes generate \textit{repeated phrases without clear meaning}, and the PPL would be extremely low with naively repeated phrases.
To measure this, the auto-BLEU score ($a$) calculates the n-gram counting within the sentence. Given text sentence $t$ and the set of n-gram $NG(t)$, auto-BLEU score of sentence $t$ is $a_t = \frac{\sum_{s \in NG(t)} \mathbf{1}[s \in (NG(t)\setminus s)]} {|NG(t)|}$. 2-gram is used for auto-BLEU calculation~\citep{gslm}. 

For the PPL of $N$ continuations ($PPL_N$), we first filter out the auto-BLEU $a_i$ higher than $\delta$, then select the lowest PPL sample as $y_c$. The threshold of the auto-BLEU score is selected by the score distribution between ground truth continuation and the generated result (Please see the Appendix \ref{appendix:autobleu} for details).
The $y_r$ is the continuation with the highest PPL. The $(y_c, y_r)$ is created as below:
\begin{equation}
  y_i = \begin{cases}
    y_c & \text{if } PPL_i = \mathbf{min}(PPL_N) \cap a_i \leq \delta \\
    y_r & \text{if } PPL_i = \mathbf{max}(PPL_N)
  \end{cases}
\end{equation} 

\subsubsection{LLM Evaluation: Mistral Score}
Instruction-tuned LLMs can follow instructions and understand semantics well~\citep{instruction-llm}. 
We propose using an LLM to judge the quality of the speech continuation, which evaluates the entire input and predicts the score. The prompt (see the Appendix for more details) is utilized to instruct the model to provide a score between 1 to 5 (1 denoting bad and 5 denoting good) based on the likelihood and meaningfulness of continuation given text prompt. Since we use the Mistral model for LLM evaluation, we call this ``\textbf{Mistral score}", denoted as $s$. To let the model learn to distinguish the preferred and unpreferred continuations, a certain threshold is set for the Mistral score to ensure the difference in quality. $s_c$ is the threshold of the chosen sample, and $s_r$ is the threshold of the rejected sample. $s_c$ should be larger than $s_r$. The auto-BLEU threshold is also used to recognize the naively repeated samples as rejected. We select the $(y_c, y_r)$ as below:
\begin{equation}
  y_i = \begin{cases}
    y_c & \text{if } s_i \geq s_c \cap a_i \leq \delta \\
    y_r & \text{if } s_i \leq s_r \cup a_i > \delta
  \end{cases}
\end{equation}
The $s_c$ and $s_r$ values are selected based on a preliminary analysis of the SLM's score distribution. 
For more details on the distribution of Mistral scores, please see Appendix \ref{appendix:score_hist}. We provide an example to illustrate the preference data selection in the Appendix \ref{appendix:mistral_example} section.

\begin{table*}[t]
\centering
\adjustbox{width=1\textwidth}{
\begin{tabular}{lccccc}
\toprule
\textbf{Evaluation} & \textbf{Metrics} & \textbf{Pre-trained} & \textbf{Fine-tuned (Diff)} & \textbf{Align-SLM w/PPL (Diff)} & \textbf{Align-SLM w/Mistral score (Diff)} \\ \hline
\multirow{3}{*}{\textbf{\shortstack[c]{Proxy Metric}}} 
& auto-BLEU $\downarrow$ & 2.80 & 2.43 (\textcolor{darkgreen}{-0.37}) & 2.58 (\textcolor{darkgreen}{-0.22}) & \textbf{2.20} (\textcolor{darkgreen}{-0.60}) \\
& PPL $\downarrow$ & 114.9 &   116.3 (\textcolor{red}{+1.4})&  \textbf{52.1} (\textcolor{darkgreen}{-62.8}) & 100.5 (\textcolor{darkgreen}{-14.4}) \\
& Mistral score $\uparrow$ & 1.66 & 1.70 (\textcolor{darkgreen}{+0.04}) & 1.88 (\textcolor{darkgreen}{+0.22}) & \textbf{2.17} (\textcolor{darkgreen}{+0.51}) \\ \hdashline

\multirow{2}{*}{\textbf{\shortstack[c]{Zerospeech}}} 
& sBLIMP $\uparrow$ & 56.8 & 56.9 (\textcolor{darkgreen}{+0.1}) & \textbf{58.9} (\textcolor{darkgreen}{+2.1}) & 58.1 (\textcolor{darkgreen}{+1.3}) \\ 
& sWUGGY $\uparrow$ & 71.8 & 70.9 (\textcolor{red}{-0.9}) & 71.9 (\textcolor{darkgreen}{+0.1}) & \textbf{72.2} (\textcolor{darkgreen}{+0.4}) \\ \hdashline
\multirow{2}{*}{\textbf{\shortstack[c]{StoryCloze}}} 

& S-StoryCloze $\uparrow$ & 52.7 & 53.0 (\textcolor{darkgreen}{+0.3}) & 52.6 (\textcolor{red}{-0.1})& \textbf{54.3} (\textcolor{darkgreen}{+1.6}) \\
& T-StoryCloze $\uparrow$ & 69.7 & 70.7 (\textcolor{darkgreen}{+1.0}) & 67.7 (\textcolor{red}{-2.0})& \textbf{74.2} (\textcolor{darkgreen}{+4.5}) \\ \hdashline

\multirow{2}{*}{\textbf{\shortstack[c]{Continuation}}} 
& GPT4-o $\uparrow$ & 1.82 & 1.83 (\textcolor{darkgreen}{+0.01}) & 1.85 (\textcolor{darkgreen}{+0.03}) & \textbf{2.06} (\textcolor{darkgreen}{+0.24}) \\
& MOSnet $\uparrow$ & 3.99 & \textbf{4.08} (\textcolor{darkgreen}{+0.09}) & 4.00 (\textcolor{darkgreen}{+0.01}) & 3.98 (\textcolor{red}{-0.01}) \\ \bottomrule
\end{tabular}}
\caption{Comparison of pre-trained model, fine-tuned model with next token prediction, Align-SLM with PPL, and Align-SLM with Mistral score on Zerospeech, StoryCloze, and speech continuation task using Librispeech test-clean. ``Diff" means the value difference compared to the Pre-trained model. The number in \textcolor{darkgreen}{darkgreen} indicates improvement over pre-trained models' performance, while \textcolor{red}{red} color stands for performance degradation.}
\label{tab:PPL_mistral}
\end{table*}

\subsection{Direct Preference Optimization for SLMs}
\label{sec:dpo}
In our framework, training with online metrics calculation is computationally infeasible due to the chain of models involved (vocoder, ASR, and LLM evaluator) and the computational complexity associated with sampling the SLM multiple times. Instead of calculating the reward online like RLHF, we adopt DPO, a simplified version of RLHF with implicit reward modeling, for preference optimization. The preference data pairs can be prepared offline, making the training more efficient. Additionally, DPO training is stable, simple, and does not require training an external reward model. The DPO training objective is 
\begin{equation}
\begin{split}
\mathcal{L}_{DPO} = 
& -\mathbb{E}_{(x,y_c,y_r)\sim D} \left[ \log \sigma \left( \beta \log \frac{\pi_\theta(y_c | x)}{\pi_{ref}(y_c | x)} \right.\right. \\
& \left. \left.-\beta \log \frac{\pi_\theta(y_r | x)}{\pi_{ref}(y_r | x)} \right) \right]
\end{split}
\end{equation}
where $\pi_{ref}$ is the reference model with the pre-trained model's parameters, which is kept frozen. $\pi_\theta$ is the policy model trained with LoRA~\citep{lora} adapter, while the parameters of the backbone pre-trained model are fixed. $\beta$ controls the deviation from the reference model $\pi_{ref}$.

\subsection{Coupling with Curriculum Learning}
\label{sec:cl}
Curriculum Learning (CL) is a machine learning approach where models are trained by gradually increasing the complexity of tasks, allowing them to learn simpler concepts first before tackling more difficult ones~\citep{cl}.
In this work, we propose to couple DPO with curriculum learning to iteratively improve automated preference data selection. We iteratively raise the difficulty in discerning the preference data by tuning the thresholds $s_c$ and $s_r$ in Equation 2. Specifically for the Mistral score, we raise the $s_c$ from 3 to 4 for chosen samples and $s_r$ from 1 to 2 for rejected samples. With Curriculum learning, we expect the model to iteratively improve, given better feedback data.

\section{Experiments}

\subsection{Dataset}
\label{sec:data}
We use \textbf{LibriSpeech}~\citep{librispeech} as our primary dataset. We use the official training, dev-clean, and test-clean set. To further expand our dataset size, we leverage the English subset of the \textbf{Multilingual Librispeech (MLS)}~\citep{mls} as an additional training set, which is \textit{around 3 times larger than the Librispeech training set}. We use a subset of the MLS data comprising 673K utterances in this work for data scaling, denoted as \textbf{\textit{mls}}. We apply the following data pre-processing steps to create the final training data:\\
\textbf{1) Speech prompt segment selection using word alignment:} Since our task involves speech continuation, the speech prompt should contain a sufficient amount of contextual information. We filter out samples shorter than 6 seconds. Unlike previous works that directly split the first 3 seconds as the prompt~\cite{gslm, twist}, we use forced alignment to select the closest word boundary around 3 seconds. This \textit{avoids cutting off spoken words in the middle}, which could cause ASR errors in the speech continuations generated by the model. This potentially leads to poor perplexity or LLM evaluation scores. \\
\textbf{2) Filtering out unsuitable chosen/rejected pairs:} We apply a second layer of filtering over Mistral score annotations of ASR transcripts by thresholding chosen and rejection scores to ensure separability. Some samples fail to create preference data pairs due to these thresholds. When multiple continuations have the same lowest or highest score, we on-the-fly randomly choose among them. The number of preference data samples for different setups is listed in Table \ref{tab:data} in the Appendix.

\begin{table*}[t]
\centering
\adjustbox{width=1\textwidth}{
\begin{tabular}{llccccccc}
\toprule
\multirow{2}{*}{\textbf{\#}} & \multirow{2}{*}{\textbf{Method}} & \multicolumn{2}{c}{\textbf{Zerospeech}} & \multicolumn{2}{c}{\textbf{StoryCloze}} & \multicolumn{3}{c}{\textbf{Speech Continuation}} \\ \cmidrule(lr){3-4} \cmidrule(lr){5-6} \cmidrule(lr){7-9}
 &  & \textbf{sBLIMP$\uparrow$} & \textbf{sWUGGY$\uparrow$} & \textbf{S-StoryCloze$\uparrow$} & \textbf{T-StoryCloze$\uparrow$} & \textbf{Mistral$\uparrow$} & \textbf{GPT4-o$\uparrow$} & \textbf{MOSnet$\uparrow$} \\ \hline
\multicolumn{9}{l}{\textit{\textbf{\textless 1B}}}                                                                                          \\
1 & GSLM~\citep{gslm}            & 54.2             & 64.8             & 53.3                & 66.6                  &$\emptyset$           &$\emptyset$         &$\emptyset$         \\
2 & AudioLM~\citep{audiolm}        & \textbf{64.7}             & 71.5             &$\emptyset$                  &$\emptyset$                    &$\emptyset$           &$\emptyset$         &$\emptyset$         \\
3 & \citet{scaling}        & 61.3             &$\emptyset$            & 56.7                   & 78.0                     &$\emptyset$           &$\emptyset$         &$\emptyset$         \\
4 & SyllableLM~\citep{syllableLM}        & 63.7             & 72.2            & $\emptyset$                 & 75.4                     &$\emptyset$           &$\emptyset$         &$\emptyset$         \\ \hline
\multicolumn{9}{l}{\textit{\textbf{1.3B}}}                                                                                                  \\
5 & \underline{VoxtLM}~\citep{voxtlm}          & 57.1             & 66.1             &$\emptyset$                  &$\emptyset$                    &$\emptyset$           &$\emptyset$         &$\emptyset$         \\
6 & TWIST~\citep{twist}           & 57.0             & 72.7             & 52.4                & 70.6                  &$\emptyset$           &$\emptyset$         &$\emptyset$         \\
7 & Pre-trained TWIST*    & 56.8             & 71.8             & 52.7                & 69.7                  & 1.66         & 1.82       & 3.99       \\
8 & Align-SLM       & 58.1             & 72.2             & 54.3                & 74.2                  & 2.17         & 2.06       & 3.98       \\
9 & Align-SLM + CL        & 58.2             & 72.2             & 53.9                & 76.1                  & 2.35         & 2.29       & 4.05 \\
10 & Align-SLM-\textit{mls}       & 59.0             & 72.7    & 54.0                & 76.7                  & 2.37         & 2.34       & 4.00       \\
11 & Align-SLM-\textit{mls} + CL        & 59.8    & 72.7    &  55.0       & 80.0        & 2.50 & 2.43 & 3.94       \\\hline
\multicolumn{9}{l}{\textit{\textbf{7B}}}                                                                                                    \\
12 & \underline{Moshi}~\citep{moshi}       & 58.8             & 72.6             & 60.8               & 83.0                  & $\emptyset$         &$\emptyset$      &$\emptyset$      \\
13 & \underline{SPIRIT-LM}~\citep{spiritlm}       & 58.3             & 69.0             & 61.0                & 82.9                  & $\emptyset$         &$\emptyset$      &$\emptyset$      \\
14 & TWIST~\citep{twist}           & 59.0             & 73.9             & 55.3                & 74.1                  &$\emptyset$           &$\emptyset$         & $\emptyset$          \\
15 & Pre-trained TWIST*    & 58.8             & 73.5             & 55.1                & 75.4                &  2.03         & 2.70       & 3.80       \\
16 & Align-SLM       & 61.1             & 75.3             & 59.1                & 83.8                  & 2.89         & 3.50       & 4.08       \\
17 & Align-SLM + CL       & 61.4             & 75.5             & 58.2                & 85.6                  & \textbf{3.22} & \textbf{3.56} & \textbf{4.09} \\
18 & Align-SLM-\textit{mls}       & 62.2             & 77.5             & 58.6                & 85.6                  & 2.92         & 3.46       & 4.02       \\
19 & Align-SLM-\textit{mls} + CL       & 62.3    & \textbf{77.9}    & \textbf{61.1}       & \textbf{86.8}         & 3.11         & 3.50       & 3.99       \\\hline
\multicolumn{9}{l}{\textit{\textbf{Cascade Topline and Human Performance}}}   \\
20 & ASR+LLM~\citep{spiritlm}       & 71.6             & 79.2             & 75.7                & 94.8                  &$\emptyset$           &$\emptyset$         &$\emptyset$         \\
21 & Human~\citep{twist}       &$\emptyset$               &$\emptyset$               & 79.2                & 90.2                  &$\emptyset$           &$\emptyset$         &$\emptyset$         \\
22 & Ground Truth Continuation       &$\emptyset$               &$\emptyset$               &$\emptyset$               &$\emptyset$                 & 2.89           &  4.25         & 4.02          \\

\bottomrule                   
\end{tabular}}
\caption{Performance on Zerospeech, StoryCloze, and Speech Continuation task. The methods marked with \underline{underline} indicate they are trained with paired speech and text tokens, \textit{not the speech-only model}. ``Pre-trained TWIST*" uses open-sourced TWIST model checkpoint but we use \texttt{bf16} precision for inference, so the performance is slightly different compared to published ``TWIST" results. ``ASR+LLM" is using Whisper~\citep{whisper} as ASR model with  Llama 2 model~\citep{llama2}, reported by \citet{spiritlm}.}

\label{tab:main}
\end{table*}

\subsection{Objective Evaluation}
\subsubsection{Zerospeech 2021 Benchmark}
sWUGGY and sBLIMP metrics evaluate SLMs' lexical and syntactic modeling on pure speech input~\cite{zerospeech}. \textbf{sWUGGY} tests if models prefer real words over phonetically similar non-words. We typically use the "in-vocab" split for reporting results, following the standard practice established by \citet{gslm}. \textbf{sBLIMP} assesses grammaticality judgments between correct and incorrect sentences. Both metrics compare geometric means of sequence probabilities assigned to paired utterances. 
\subsubsection{Spoken StoryCloze Benchmark}
StoryCloze benchmarks~\citep{storycloze} evaluate the model's ability to identify the more plausible ending among two scenarios given a short story as a prompt. This requires a degree of high-level semantic understanding and common sense. We utilize the spoken version of the original StoryCloze (\textbf{S-StoryCloze}) as well as the topic-based StoryCloze (\textbf{T-StoryCloze}) created by \citet{twist}. T-StoryCloze uses simpler negative samples that are randomly drawn while S-StoryCloze uses adversarially curated negative samples. The random baseline performance for the above tasks is 50\%. We name the spoken version of StoryCloze as ``\textbf{StoryCloze}" for simplicity, but note that this is different from the text StoryCloze.
\subsubsection{Generative Speech Continuation} 
\textbf{GPT4-o score}: GPT4-o~\citep{gpt4} has shown remarkable text understanding performance and can serve as alternative human evaluators~\citep{llm_eval, g-eval, closer}, showing high correlation with human judgments. We leverage GPT4-o as a proxy for human evaluations. Following \textit{llm evaluation}~\citep{llm_eval}, the instruction first analyzes the sentence and then provides the score from 1 to 5, to judge the semantic coherence, meaningfulness, and grammatical correctness of the ASR transcribed continuation given a prompt (1 denoting bad and 5 denoting good). The instruction prompt is shown in Appendix \ref{appendix:gpt4_prompt}. \\
\textbf{MOSnet score}: To measure the audio quality, we utilize the MOSnet~\citep{mosnet} to predict the Mean Opinion Score (MOS) of audio quality. Specifically, MOSnet is based on self-supervised wav2vec 2.0~\citep{wav2vec2}, fine-tuned on MOS prediction task\footnote{https://github.com/nii-yamagishilab/mos-finetune-ssl}. The model has shown a high correlation with human MOS scores and good generalization ability for unseen data. 

\subsection{Subjective Evaluation}
We conducted human listening evaluations to assess the meaningfulness of the generated speech. We follow \citet{gslm, pgslm} to use the \textbf{M}eaningfulness \textbf{M}ean \textbf{O}pinion \textbf{S}core (\textbf{MMOS}). 
Specifically, given a speech prompt and generated speech continuations, evaluators listen to audio samples and rate the meaningfulness in terms of relevance, coherence, and grammatical correctness. 
We randomly sample 100 speech prompts from the Librispeech test-clean set for evaluation.
Each sample has 10 evaluators to provide the rating on a scale between 1 to 5 with an increment
of 1.  
The human evaluation template and instruction are shown in Appendix \ref{appendix:human}. 
CrowdMOS~\citep{crowdmos} package is used for outlier removal~\citep{gslm}.  

\subsection{Baselines}
We compare Align-SLM with other SLMs on the Zerospeech and StoryCloze benchmarks. The most comparable baselines are \textit{speech-only SLMs}, including GSLM~\citep{gslm}, AudioLM~\citep{audiolm}, TWIST~\citep{twist}, SyllableLM~\citep{syllableLM}, and the model from \citet{scaling}. Among these, the TWIST model is initialized from a text-based LLM.
Additionally, we compare the performance against \textit{SLMs leverage text modality} (Table \ref{tab:main} with \underline{underline}), specifically VoxtLM~\citep{voxtlm} with multi-task training, SPIRIT-LM~\citep{spiritlm} with speech-text interleaving, and Moshi~\citep{moshi}, which leverages text-guided speech generation.

\section{Results}

\begin{table}[]
\centering
\adjustbox{width=0.4\textwidth}{
\begin{tabular}{lc}
\toprule
\textbf{Method}                & \textbf{MMOS$\uparrow$}        \\ \hline
Target Re-synthesized & 3.50 ± 0.07 \\
Pre-trained TWIST 7B  & 3.48 ± 0.07 \\
Align-SLM 7B + CL     & \textbf{3.73 ± 0.06} \\ \bottomrule
\end{tabular}}
\caption{Meaningfulness MOS score. We report the MMOS score as the mean ± 95\% confidence interval of the standard deviation. ``Target Re-synthesized" uses the speech tokens from the original continuation and re-synthesize back to the waveform.}
\label{tab:mmos}
\end{table}

\subsection{Mistral Score Provides Better Semantic Feedback Than Perplexity}
\label{sec:mistral_better}
To determine which preference data selection strategy is beneficial for SLM's semantics, we first conduct preliminary experiments on Align-SLM with PPL and Mistral score in Table \ref{tab:PPL_mistral} using the TWIST 1.3B model. Additionally, we continually fine-tune the pre-trained model using the same data, which serves as the baseline for the next speech token prediction. 
"Proxy Metric" refers to the metrics used for preference data selection, while ``Zerospeech", ``StoryCloze", and ``Speech Continuation" are zero-shot speech evaluation metrics.

Align-SLM w/PPL successfully improves the proxy metrics for auto-BLEU and PPL, but the performance on speech continuation is slightly worse than the pre-trained model. 
As for the Zerospeech and StoryCloze benchmark, Align-SLM w/PPL has marginal improvement on most metrics and degrades on T-StoryCloze. 
Particularly, perplexity feedback shows a much greater improvement on sBLIMP, which measures grammatical correctness. This finding suggests that optimizing toward perplexity might overly focus on grammar rather than general semantics and relevance.

On the other hand, \textit{Align-SLM w/Mistral score significantly outperforms the pre-trained model across metrics}. Specifically, the performance on the Zerospeech and StoryCloze benchmarks improve significantly  (+1.6 on S-StoryCloze and +4.5 on T-StoryCloze). The generated continuations also yield a better GPT4-o score, indicating the generated content is more relevant and coherent to the speech prompt. Regarding proxy metrics, the auto-BLEU and Mistral scores are improved, whereas the PPL is similar to the pre-trained model. 

This finding suggests that LLM evaluations, such as the Mistral score, provide general semantic feedback and achieve superior performance across benchmarks. Furthermore, Align-SLM w/Mistral score significantly outperforms the fine-tuned baseline, which only marginally improves performance. Therefore, \textit{in the following experiments for Align-SLM, we use the \textbf{Mistral score as the AI feedback}.}

\subsection{Consistently Improves Pre-trained SLMs}
Given our findings from Section \ref{sec:mistral_better}, we use the Mistral score as a ``proxy metric" for the rest of our experiments.
In Table \ref{tab:main}, we observe that Align-SLM consistently outperforms the pre-trained model on the Zerospeech, StoryCloze, and speech continuation task for both the 1.3B and 7B models (rows 7 to 8 for 1.3B, row 15 to 16 for 7B). For instance, on T-StoryCloze, training Align-SLM with Librispeech data yields relative improvements of 6.5\% and 11.1\% for the 1.3B and 7B models, respectively. 
Additionally, Table \ref{tab:main} demonstrates Align-SLM 7B performs better in speech continuation, as reflected by the GPT-4 score improving from 2.70 to 3.50.

\subsection{Improvement of Curriculum Learning}
After training the pre-trained model with the first iteration of Align-SLM, we consider the resulting model a stronger starting point and generate new preference data with more stringent preference data selection criteria. 
Results in Table \ref{tab:main} indicate that curriculum learning improves most metrics on the Zerospeech and StoryCloze benchmarks (rows 8 to 9 for 1.3B, rows 16 to 17 for 7B), particularly for T-StoryCloze, which requires fine-grained relevance between the speech story prompt and its continuation.
Additionally, we observe improvements in speech continuation; for example, the GPT4-o score increases from 2.06 to 2.29 for the 1.3B model. 
We also experimented with further increasing the number of curriculum learning iterations, which continued to enhance performance (see the discussion in Appendix \ref{appendix:iteration}). 

\subsection{Amount of Preference Data}
In the previous experiments, we only use the LibriSpeech training data for DPO training. 
After applying filtering as described in Section \ref{sec:data}, the number of samples is around 39K and 63K for the 1.3B model and 7B model, respectively. 
To investigate whether additional data helps the Align-SLM to learn semantics, we scale up the data around three times by including a subset of MLS (\textit{mls}). 
It is worth noting that the scale of the MLS subset is still much smaller than the pre-training data used in \citet{twist}. 
Table \ref{tab:main} shows that that with more data, the \textbf{model learns significantly better semantics across model sizes and benchmarks} (row 10 to 11 for 1.3B, row 18 to 19 for 7B). Nevertheless, we observe that adding \textit{mls} data for the 7B SLM yields little improvement in GPT4-o score compared to 1.3B model\footnote{This can be attributed to the amount of preference data for the 7B model from Librispeech already being sufficient (63K for the first iteration and 71K for the second iteration). In contrast, the 1.3B model only has 39K and 20K preference data pairs by Librispeech.}. 


\subsection{Comparison with Baselines}
Align-SLM-\textit{mls}-CL achieves \textbf{state-of-the-art performance} for textless SLMs in T-StoryCloze (\textbf{86.8}), S-StoryCloze (\textbf{61.1}), and sWUGGY (\textbf{77.9}), \textbf{even surpassing \textit{text-guided approaches} (moshi) and \textit{speech-text interleaving methods} (SPIRIT-LM)}. The performance on T-StoryCloze is close to human-level accuracy (90.2). However, for sBLIMP (grammatical correctness), AudioLM and SyllableLM achieve the best performance, possibly due to their superior design of speech tokens. Compared to the cascaded topline (ASR+LLM), end-to-end SLMs still have room for improvement.

\subsection{Human Prefer Align-SLM's Generation}
Table \ref{tab:mmos} presents the MMOS scores from the subjective evaluation. We compare the re-synthesized speech of the original continuation, the pre-trained TWIST 7B, and the proposed Align-SLM 7B with CL. The results show that human evaluators perceive Align-SLM as generating more meaningful speech continuations than the pre-trained model, and even surpassing the original continuation. This can be attributed to the fact that the original continuation, derived from audiobook content, may rely on the broader context, whereas Align-SLM learns to generate more relevant and meaningful content based on the speech prompt.

\subsection{Impact on Audio Quality}
Preference optimization can potentially adversarially exploit the ASR and LLM evaluators, leading the SLM to generate nonsensical or noisy speech with artificially high rewards. To address this concern, we evaluate the audio quality of generated speech using MOSnet \cite{mosnet}, which has shown a high correlation with human judgments. Table \ref{tab:main} presents the MOS scores from LibriSpeech dev-clean and test-clean sets. Results indicate that the MOSnet scores for Align-SLM are comparable to or slightly higher than those of the pre-trained SLM, suggesting that the training of Align-SLM preserves audio quality.
The proposed framework requires the generated speech to pass through the ASR model for speech-to-text conversion, natural and clear speech is essential to avoid speech recognition errors. Consequently, in some cases, the MOSnet score for Align-SLM is even higher than that of the pre-trained SLM.

\section{Conclusion}
This work introduces \textbf{Align-SLM}, a novel framework that significantly enhances the semantics of SLM via preference optimization. By utilizing LLM-guided semantic feedback and direct preference optimization, Align-SLM achieves state-of-the-art performance of SLMs across various benchmarks and generative tasks, consistently outperforming pre-trained SLMs. The framework demonstrates superior results with the proposed LLM evaluation feedback and curriculum learning. This work highlights the critical role of preference optimization for SLM and paves the way for better end-to-end speech-to-speech models.

\section*{Limitations and Future Works}
\textbf{Limited to Semantics of SLMs}: This work investigates only the semantic aspect of SLMs, though other aspects such as speaking styles, paralinguistics, and prosody are also important for spoken dialogue~\citep{paralingpt, styletalk}. Our Align-SLM framework can be generalized to these aspects, but that is beyond the scope of this paper.\\
\textbf{Integrating Align-SLM with other SLMs}: We integrated the open-source pre-trained TWIST~\citep{twist} model with Align-SLM, demonstrating significant performance improvements. As the SLM community grows rapidly, more powerful pre-trained SLMs are emerging~\citep{moshi, syllableLM, llama-omni}. The Align-SLM framework is a general framework that can be extended to integrate with other SLMs.\\
\textbf{Language Support}: The current model is trained solely on English speech data. Future work can extend the Align-SLM framework to multilingual speech data for more inclusive speech technology. For unwritten languages, instead of using ASR models for text transcription, we can use speech translation models to convert unwritten spoken languages into high-resource languages~\citep{s2s_unwritten}, similarly obtaining semantic feedback from the LLM.\\
\textbf{Diverse Data and Model Scaling}: 
The dataset used in this work is much smaller than typical pre-training datasets, and the model size is relatively small compared to text-based LLMs. Additionally, the training data is limited to the audiobook domain. Expanding the training data to include more diverse domains in future work could lead to improved model performance.

\bibliography{custom}
\appendix

\section*{Appendix}

\section{Implementation Details}
For sampling the speech continuations, we use a decoding temperature of 0.8, following the original sampling setup in \citet{twist}. 
The number of sampled continuations, $N$, is set to 5.
The hyperparameters of the LoRA adapter are set with a rank of 32 and an alpha value of 8.
For training on LibriSpeech, models are trained with a batch size of 512, using a peak learning rate of 1e-6 that linearly decays over 100k steps with 500 warmup steps. For \textit{mls} data, since there are more training samples, we further increase the training steps to 300k. 
For curriculum learning, the peak learning rate is 1e-6 and linearly decays without the warmup step.
We utilize 64 NVIDIA A100 GPUs for training. The model checkpoint is selected by the highest reward accuracy on the Librispeech dev-clean set.

\section{Details of Subjective Evaluation}
\label{appendix:human}
We used the Amazon Mechanical Turk crowdsourcing platform for subjective evaluations. The evaluators, all based in the United States, had HIT approval rates above 98\%, as the speech samples were in American English. Each evaluation task contained 10 samples, and annotators were paid 0.50 USD per task. On average, each task took about 2 minutes to complete.

\section{Preference-based selection illustration}
\label{appendix:mistral_example}
Given the speech prompt ``\textit{\textbf{a man was looking in from the corridor behind}}",
we draw five candidate continuations from the model and assign each a Mistral score
(higher is better).
Table~\ref{tab:pref-example} lists the continuations and their scores.
We keep a continuation if \(s_i \ge s_c=3\), reject it if
\(s_i \le s_r=1\), and otherwise mark it as \textit{filtered} (neither accepted nor explicitly rejected).
In this example, continuations~(1) and~(5) are kept;
continuations~(2) and~(4) are rejected due to clear
irrelevance; and continuation~(3) is filtered.
Notably, the lowest-scoring continuations (score 1) tend to be
semantically unrelated to the prompt, underscoring the value of the LLM feedback.

\begin{table}[t]
  \centering
  \footnotesize
  \setlength{\tabcolsep}{6pt}
  \begin{tabularx}{\linewidth}{@{}lXc@{}}
    \multicolumn{3}{@{}l@{}}{\textbf{Prompt:} \emph{a man was looking in from the corridor behind}}\\[4pt] \toprule
    \# & \textbf{Sampled speech continuation} & \textbf{Score} \\ \midrule
    1 & He was seen, he stopped before the man. He said, \emph{I have a book you read}. & 3 \\
    2 & When they drew near and received their letter, and looking down, showed them something of the startling effect of the diamond. & 1 \\
    3 & But every time he reached the open window he saw the little man pressing in a silk~sash. & 2 \\
    4 & When he heard the dog talking to him. Said, \emph{My father, did you hear the dog?} Said he, \emph{did you\dots} & 1 \\
    5 & And the door opened, and he heard the words. And there they all began, Boyce said to himself. & 3 \\ \bottomrule
  \end{tabularx}
  \caption{Prompt, sampled continuations, and their \textsc{Mistral} scores.}
  \label{tab:pref-example}
\end{table}

\section{Speech Continuation Examples}
We present some qualitative examples of speech continuations in Table \ref{tab:example}. We observe that the pre-trained TWIST 7B model occasionally generates nonsensical speech based on the given speech prompt. For instance, in the first example, the TWIST continuation includes the repetitive phrase "only, easy, only, and" at the end.

\begin{table*}[ht]
\centering
\renewcommand{\arraystretch}{1.2} 
\adjustbox{width=0.95\textwidth}{
\begin{tabular}{p{5cm}p{6cm}p{6cm}}
\toprule
\textbf{Speech Prompt} & \textbf{Pretrained TWIST Continuation} & \textbf{Align-SLM Continuation} \\
\hline
He came down to earth lived among men suffered & 
for a friend, she would have been glad of the last word, the pointed word, possible only, easy, only, and. & 
died, and yet somehow he felt that he lived more than they, and he came to know men of his own who loved him still. \\
\hline
Then there were three or four leading men of the community & 
and the majority of these were hirelings, hard-headed and stolid. & 
with them at the table, and they spoke together about the measures they were preparing for the day's work. But one among them was speaking feelingly. \\
\hline
In the early morning of the next day may twenty ninth & 
W. H. Barker, the prisoner before the magistrate and the charge of the prosecution, & 
we went to our rooms and sat down and had a long talk over all the present circumstances of our journey to the place where we had spent so much time. \\
\bottomrule
\end{tabular}}
\caption{Continuation examples from Pretrained TWIST 7B model and Align-SLM 7B + CL. The transcribed texts are shown in the table.}
\label{tab:example}
\end{table*}

\section{Number of Iteration for Curriculum Learning}
\label{appendix:iteration}
\begin{figure*}[t]
  \centering
\includegraphics[width=0.8\linewidth]{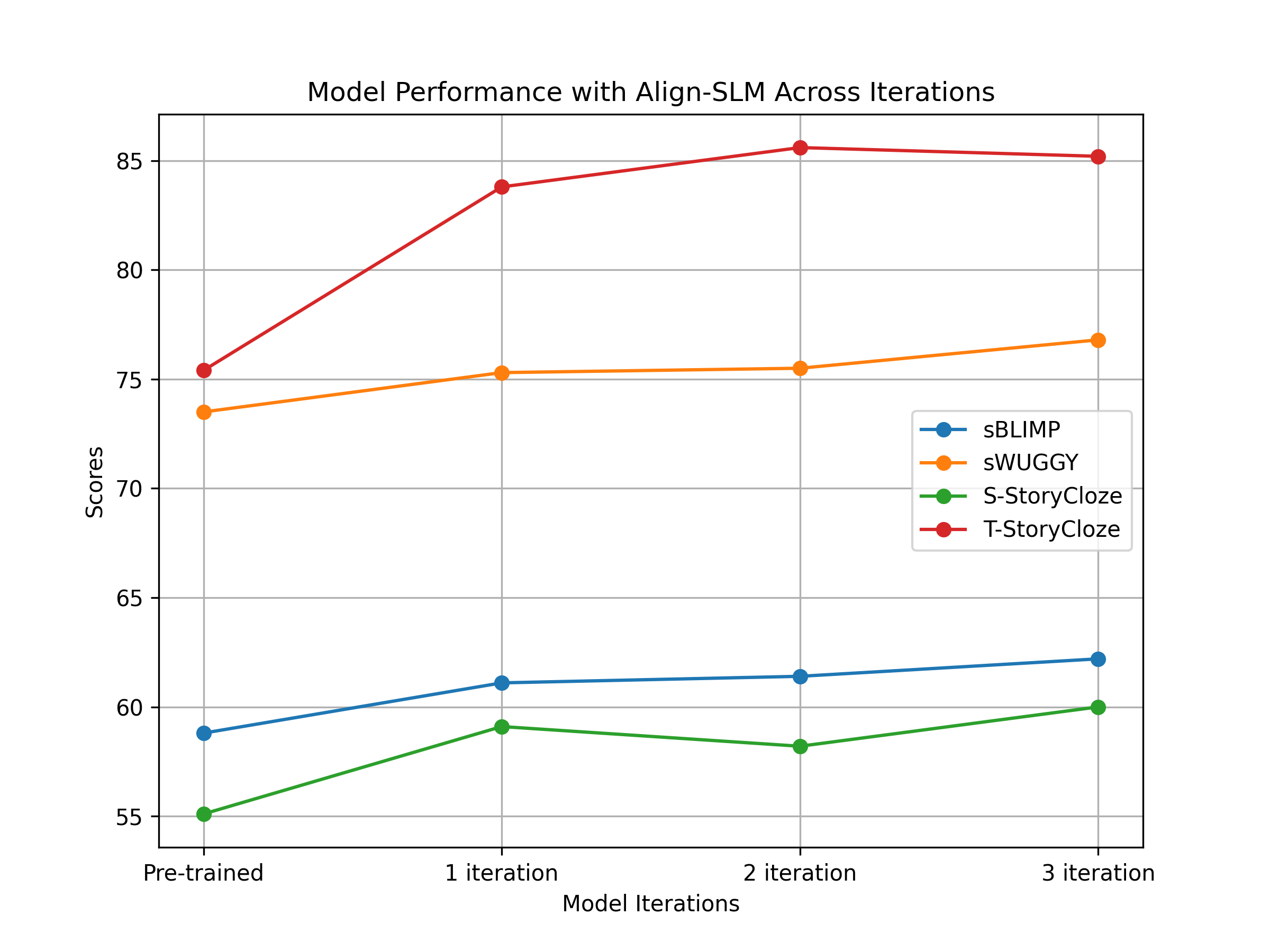}
  \caption{The 7B model undergoes one, two, and up to three iterations of curriculum learning with Align-SLM using Librispeech data.}
  \label{fig:iteration}
\end{figure*}

In this section, we examine how the number of curriculum learning iterations influences performance. Figure \ref{fig:iteration} presents the results from the pre-trained 7B model after one to three iterations of Align-SLM with curriculum learning using Librispeech data. For the first, second, and third iterations, $(s_r, s_c)$ is $(1, 3)$, $(2, 4)$, and $(3, 5)$, respectively. We observe that increasing the number of curriculum learning iterations gradually improves the SLM's performance across all benchmarks. While the T-StoryCloze performance saturates after the second iteration, other metrics show their best performance in the third iteration.

\section{Instruction for the Mistral LLM for evaluation}
The [text\_prompt] and [generated\_transcription] are the placeholder for different input. The entire instruction is shown as following: \\
\texttt{Rate the text continuation based on how likelihood is the text continuation given the text prompt. You should also consider whether the meaning of the text continuation is making sense. Don't be too strict. The text prompt is [text\_prompt], and the text continuation is [generated\_transcription].\\
You must give an overall rating from 1 to 5. The rating guideline is \\
1: Very Unlikely and Irrelevant; \\
2: Unlikely and Marginally Relevant; \\
3: Moderately Likely and Relevant; \\
4: Likely and Relevant; \\
5: Very Likely and Highly Relevant. \\
Output format is: I would rate the score as [NUMBER]}

\section{Instruction for the GPT4-o for evaluation}
\label{appendix:gpt4_prompt}
\texttt{The task is evaluating the relevance and likelihood of the predicted text continuation, given the text prompt. You should also consider whether the meaning of the text continuation is making sense. The text prompt is: [text\_prompt], and the text continuation is :[generated\_audio\_transcription]. \\\\
    You must give an overall rating from 1 to 5. The rating guideline is as below: \\
    1: The text continuation is very unlikely and irrelevant to the text prompt.\\
    2: The text continuation is unlikely and marginally relevant to the text prompt.\\
    3: The text continuation is moderately likely and relevant to the text prompt.\\
    4: The text continuation is likely and relevant to the text prompt.
    5: The text continuation is very likely and highly relevant. \\
    You should take the following steps to provide the score: \\
    First: briefly analyze the sample with the above definition. \\
    Second: MUST follow the output format as: I would rate the score as \_}

\section{auto-BLEU Score Distribution}
\label{appendix:autobleu}
\begin{figure*}[t]
  \centering
  \includegraphics[width=0.8\linewidth]{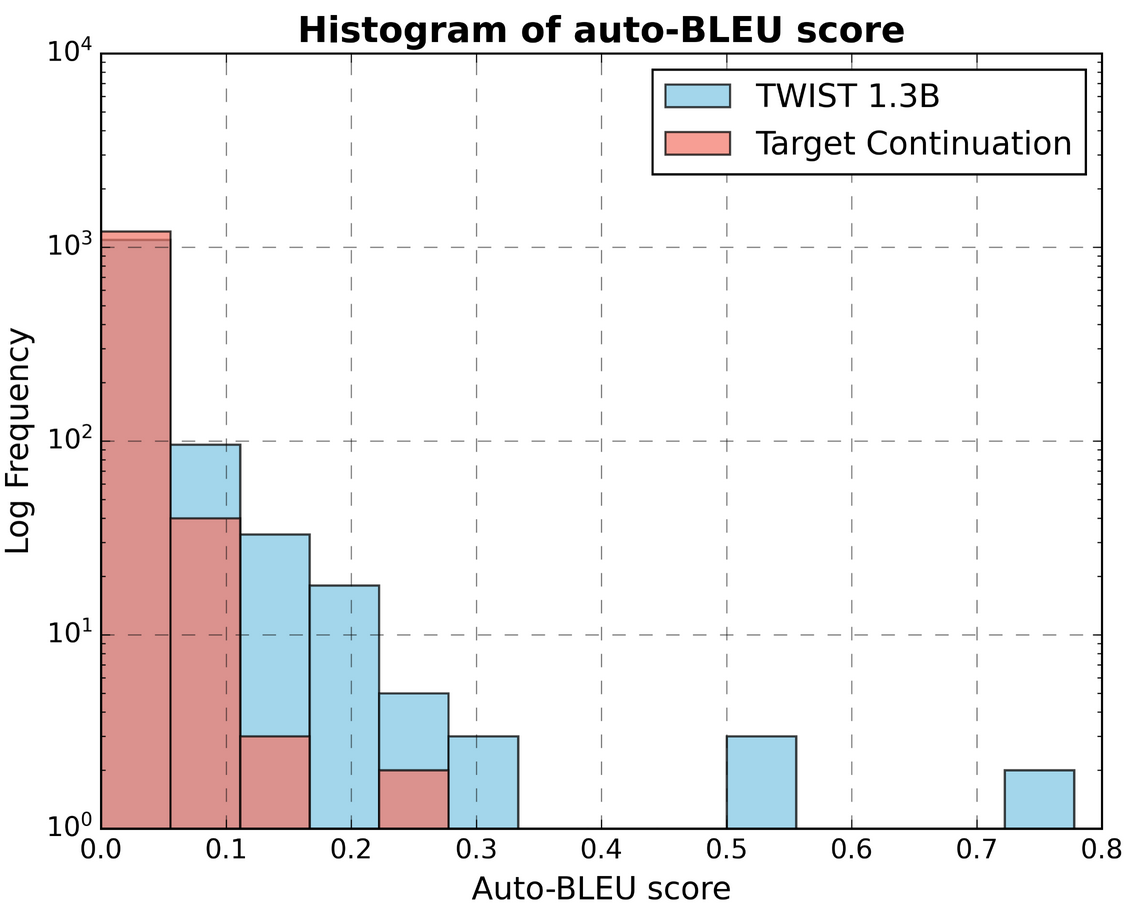}
  \caption{The log normalized histogram of auto-BLEU score distribution on Librispeech dev-clean with 1.3B TWIST model.}
  \label{fig:autobleu_dist}
\end{figure*}

Table \ref{fig:autobleu_dist} presents the auto-BLEU score distribution on of the TWIST 1.3B model and the target continuation. We observe that most target continuations have an auto-BLEU score lower than 0.1, while TWIST's generated continuations show auto-BLEU scores ranging from 0.1 to 0.3 (or even higher). This difference suggests that an auto-BLEU score higher than 0.1 may not indicate a good continuation. Therefore, we set the auto-BLEU score threshold ($\delta$) at 0.1 to distinguish the chosen and rejected continuations.

\section{Mistral Score Distribution}
\label{appendix:score_hist}
\begin{figure*}[t]
  \centering
  \includegraphics[width=0.8\linewidth]{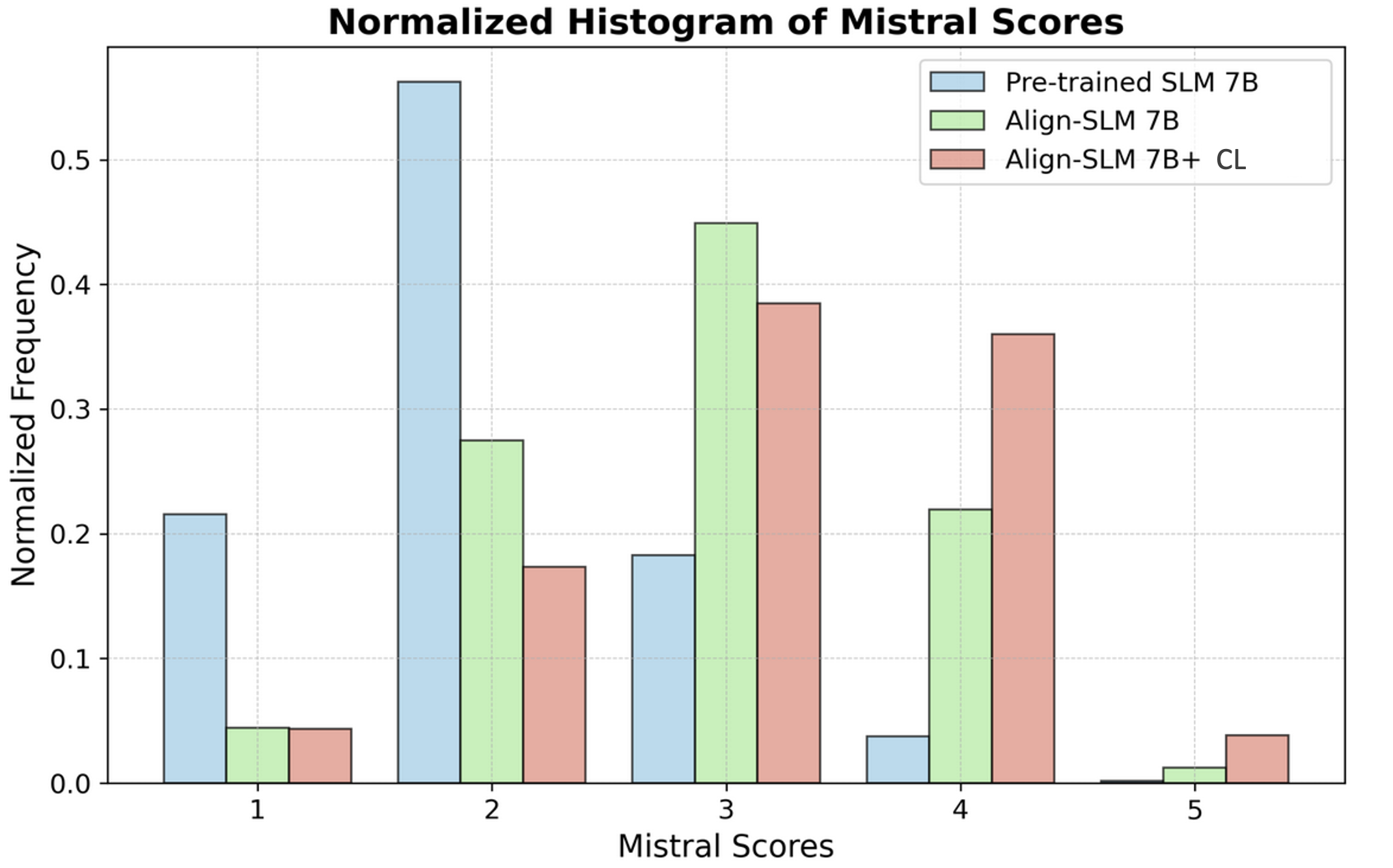}
  \caption{The normalized histogram of Mistral score distribution on Librispeech dev-clean with 7B model.}
  \label{fig:hist_7b}
\end{figure*}

\begin{figure*}[t]
  \centering
  \includegraphics[width=0.8\linewidth]{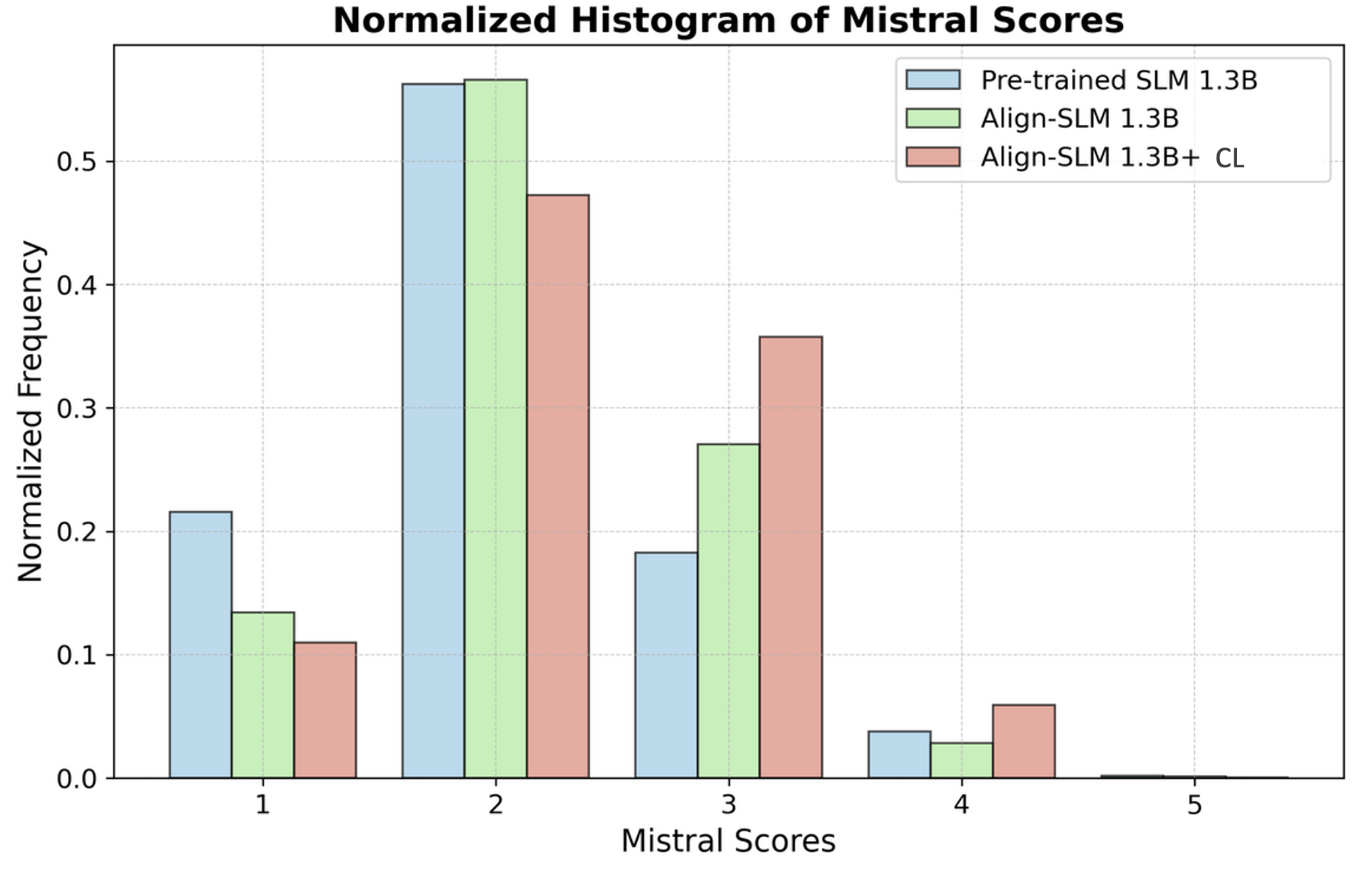} 
  \caption{The normalized histogram of Mistral score distribution on Librispeech dev-clean with 1.3B model.}
  \label{fig:hist_1.3b}
\end{figure*}

Given $N$ generated speech continuations, selecting an appropriate scoring threshold is crucial for determining which samples are chosen or rejected. Thresholds that are too high or too low fail to create effective preference data pairs. In this work, we first analyze the score distribution of the pre-trained SLMs, as shown in Figures \ref{fig:hist_7b} and \ref{fig:hist_1.3b}. We observe that the Mistral score is 2 with more than a 50\% probability for both the 1.3B and 7B pre-trained SLMs. Therefore, a score of 2 is considered the norm, a score of 1 represents rejected cases, and scores of 3 or higher are considered chosen samples.
After the initial Align-SLM DPO training, the score distribution shifts towards higher scores, reducing the proportion of low scores. Increasing the $s_c$ and $s_r$ values in the second iteration can further improve the overall ratings.

\section{Correlation Between GPT4-o score and MMOS}
\label{appendix:correlation}
We calculate the Pearson's correlation coefficient between the GPT4-o score and MMOS score. Pearson's coefficient is 0.51, suggesting that the GP4-o score correlates well with the human rating. The finding is aligned with previous works that the rating from GPT4 can be alternative human evaluators~\citep{llm_eval, g-eval, emphasis}.


\begin{table}[t]
\centering
\adjustbox{width=0.35\textwidth}{
\begin{tabular}{lcc}
\toprule
\multicolumn{1}{c}{\textbf{Dataset}} & \textbf{Iter 1} & \textbf{Iter 2} \\
\hline
\multicolumn{3}{l}{\textit{\textbf{1.3B}}}                                                                         \\
Librispeech                          & 39,566                       & 20,942                       \\
+ \textit{mls}                          & 131,071                      &   113,243                          \\
\hline
\multicolumn{3}{l}{\textit{\textbf{7B}}}                                                                           \\
Librispeech                          & 63,107                       & 71,944                       \\
+ \textit{mls}                          & 234,195                      &     262,929                        \\
\bottomrule                      
\end{tabular}}
\caption{Number of data samples after filtering. Note that different models have different filtered samples since the filtering depends on the quality of the generated speech. The number of samples of original Librispeech and \textit{mls} are 247K and 673K, respectively. }
\label{tab:data}
\end{table}

\begin{figure*}{}
    \centering
    \begin{mdframed}
\includegraphics[width=1\linewidth]{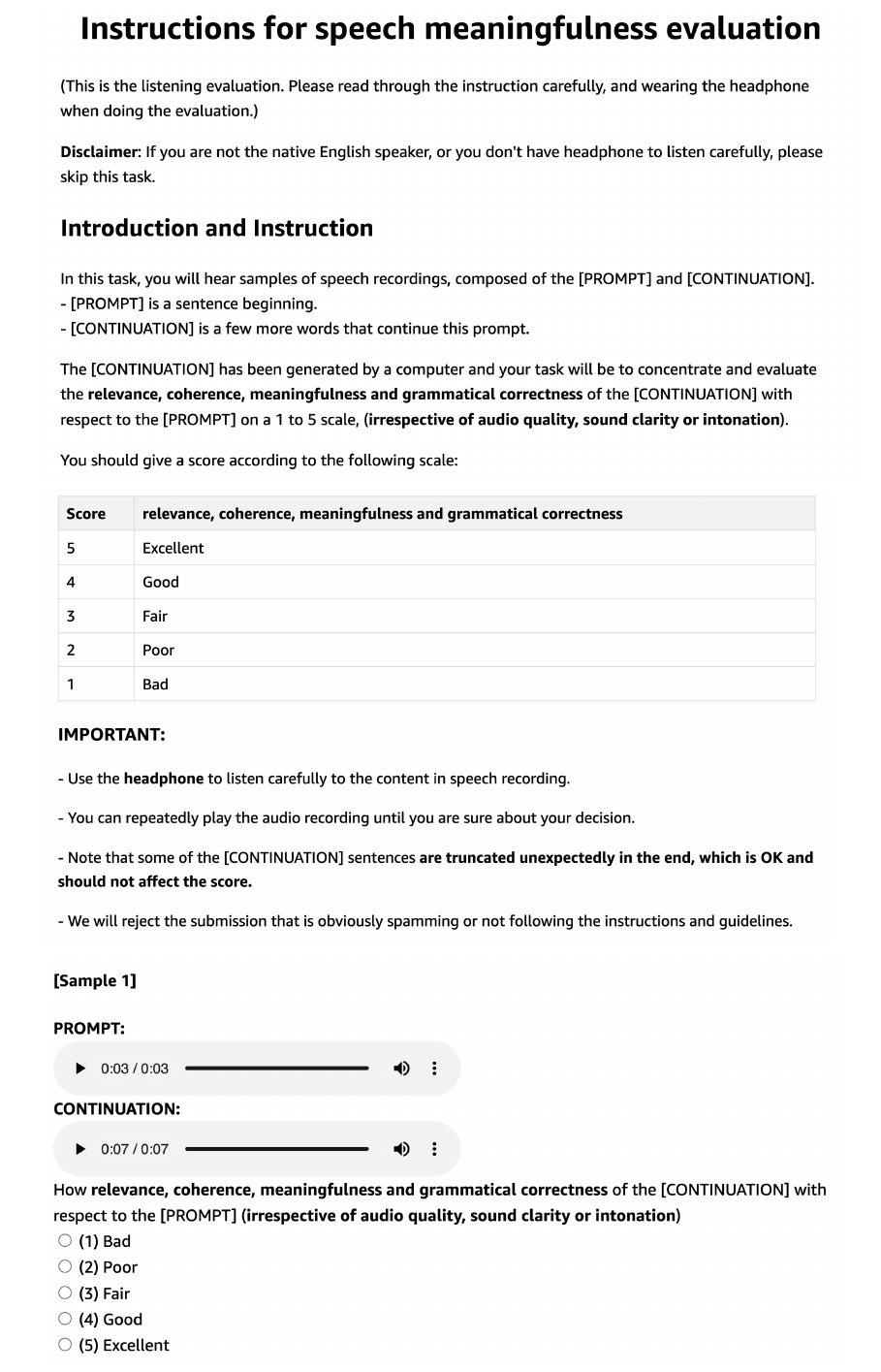}
    \end{mdframed}
    \caption{The template and instruction of the subjective evaluation of the MMOS score.}
    \label{fig:subjective_eval_template}
\end{figure*}

\end{document}